\documentclass[journal]{IEEEtran}
\usepackage{amsmath,amsfonts,amssymb}
\usepackage{algorithmic}
\usepackage{algorithm}
\usepackage{array}
\usepackage[caption=false,font=normalsize,labelfont=sf,textfont=sf]{subfig}
\usepackage{textcomp}
\usepackage{stfloats}
\usepackage{url}
\usepackage{verbatim}
\usepackage{graphicx}
\usepackage{booktabs}
\usepackage{multirow}
\usepackage{makecell}
\usepackage{soul}
\usepackage{placeins}
\usepackage{pgfplots}
\pgfplotsset{compat=1.18} 
\usepackage{graphicx}
\usepackage{subcaption}
\usepackage[table]{xcolor}
\usepackage{pdfpages} % Allows inserting external multi-page PDFs

\begin{document}

\title{Semi-Supervised Adaptation of Vision-Language Models for Image Classification}

\author{Mohamed~L.~Mekhalfi,~\IEEEmembership{Senior~Member,~IEEE,}
        Mohamad~M.~Al~Rahhal,~\IEEEmembership{Senior~Member,~IEEE,}
        Yakoub~Bazi,~\IEEEmembership{Senior~Member,~IEEE,}
        Salah~E.~Khenfer,
        Mingdeng~Shi,
        Hua~Zou, and~Mansour~Zuair,~\IEEEmembership{Member,~IEEE}%

\thanks{* Corresponding Author: mmaalrahhal@ksu.edu.sa}
\thanks{M.~L.~Mekhalfi is with Fondazione Bruno Kessler, Via Sommarive 18, 38123, Povo TN, Italy.
Mohamad~M.~Al~Rahhal is with the Applied Computer Science Department, College of Applied Computer Science, King Saud University, 11543, Riyadh, Saudi Arabia.
Y.~Bazi and M.~Zuair are with the Computer Engineering Department, College of Computer and Information Sciences, King Saud University, 11543, Riyadh, Saudi Arabia.
S.~E.~Khenfer and M.~Shi are with the Key Laboratory of Tarim Oasis Agriculture, College of Information Engineering, Tarim University, Aral, China.
H.~Zou is with the School of Computer Science, Wuhan University, 430072, Wuhan, China.}% <-this % stops a space
\thanks{ }}

\markboth{}%
{Shell \MakeLowercase{\textit{et al.}}: Bare Demo of IEEEtran.cls for Journals}

\maketitle

\begin{abstract}

Vision-language models like CLIP have shown significant potential in handling natural images, yet their performance is often limited by the distinct characteristics of satellite imagery. 
While parameter-efficient adaptation techniques exist, their efficacy is frequently limited by the scarcity of annotated samples. 
In this letter, we propose Self-Evolutionary CLIP (SE-CLIP), a semi-supervised framework designed for recursive label mining in scene classification. 
The approach follows a dual-phase pipeline, where an initial warm-up on a few annotated seeds is followed by a recursive discovery phase that iteratively identifies high-confidence samples from unlabeled pools. 
To maintain the integrity of the evolving support set, we employ a class-balanced selection strategy that prevents the model from being dominated by easily learned categories. 
Results on the UCM and NWPU benchmarks indicate that SE-CLIP significantly outperforms existing semi-supervised approaches. The framework provides a viable solution for adapting VLMs to the remote sensing domain with minimal human intervention.
\end{abstract}


\IEEEpeerreviewmaketitle

\section{Introduction}

\IEEEPARstart{T}{he} emergence of Vision-Language Models (VLMs), such as Contrastive Language-Image Pre-training (CLIP) \cite{li2026clip}, has revolutionized visual representation learning. By establishing a unified latent space, these models enable zero-shot classification in Remote Sensing (RS) through semantic alignment. However, a critical performance gap persists due to the significant domain discrepancy between general-purpose pre-training and the specialized spectral-spatial geometries of satellite imagery.

To mitigate this domain discrepancy, parameter-efficient fine-tuning techniques, such as Low-Rank Adaptation (LoRA), have been introduced \cite{lora}. By updating only a small subset of additional parameters while keeping the backbone weights frozen. These methods allow for domain specialization without the computational burden of full fine-tuning. Yet, the efficacy of such adaptation remains heavily dependent on the availability of high-quality, expert-annotated labels. In the RS domain, where data labeling requires specialized geographical knowledge and is notoriously labor-intensive, this dependence creates a critical bottleneck for large-scale monitoring applications.

Semi-supervised learning (SSL) offers a potential solution by leveraging vast amounts of unlabeled data. However, traditional SSL frameworks often struggle with the noisy nature of satellite imagery, leading to confirmation bias or the reinforcement of majority-class errors during pseudo-labeling.

Existing semi-supervised frameworks, ranging from holistic consistency regularization to curriculum-based pseudo-labeling \cite{flexmatch, msmatch, rsmatch, freematch, softmatch, fixmatch, semirs, cpl, darp}, primarily rely on heuristic or self-adaptive thresholding to filter unlabeled samples. While effective in closed-set scenarios, these methods encounter three critical bottlenecks in the RS domain. 
First, thresholding-based selection is highly sensitive to the initial model confidence. Precisely, in the presence of a domain gap, low zero-shot accuracy leads to the exclusion of informative samples or the inclusion of high-confidence errors. 
Second, the reliance on absolute thresholds in these methods introduces a significant risk of including erroneous labels. For instance, samples from easily distinguishable categories frequently exceed the confidence threshold, while complex or visually ambiguous classes are systematically excluded. This imbalance creates a self-reinforcing bias that biases the adapted feature space toward majority-class characteristics, ultimately degrading the model's sensitivity to the full diversity of land-use semantics.
Third, these works remain confined to uni-modal architectures with standard linear heads, failing to exploit the rich, high-dimensional semantic priors available in Vision-Language Models (VLMs). By ignoring the cross-modal alignment capabilities of VLMs like CLIP, traditional SSL methods overlook a robust initialization source that can guide the discovery of unlabeled samples even before extensive fine-tuning. 

Our framework addresses these gaps by replacing rigid thresholding with a recursive discovery protocol anchored in CLIP’s semantic latent space. This framework, termed SE-CLIP (Self-Evolutionary CLIP), transforms the adaptation process into a dual-phase pipeline that initiates from a sparse set of expert-labeled seeds. Unlike traditional methods that rely on static pseudo-labeling, SE-CLIP utilizes fixed textual anchors to guide the label discovery and self-training of the model. By parameterizing the similarity scaling factor as a learned multiplicative variable, the framework adaptively calibrates its alignment confidence, enabling the robust identification of high-confidence samples even under significant domain shifts. This process is governed by a class-balanced migration strategy that enforces an equidistribution of mined samples across the label space, ensuring a high-purity expansion of the support set and preventing dominance by easy classes.

The remainder of this letter is organized as follows. Section \ref{sec:method} details the SE-CLIP framework, including the architectural adaptation and the recursive discovery protocol. Section \ref{sec:experiments} presents and discusses the results. Finally, Section \ref{sec:conclusion} concludes the letter and suggests potential research directions.
\section{Methodology}
\label{sec:method}

\subsection{Problem Formulation}

The proposed SE-CLIP framework establishes a label mining pipeline designed to bridge the domain discrepancy between general-purpose vision-language models and the specific spectral-spatial requirements of remote sensing (RS) scene classification. By integrating parameter-efficient low-rank adaptation with a cross-modal alignment objective, the proposed framework enables the transition from a zero-shot foundation model to a domain-specialized classifier.

In the context of semi-supervised remote sensing imagery classification, we consider a dataset $\mathcal{D}$ partitioned into a labeled support set $\mathcal{L}$ and a significantly larger unlabeled pool $\mathcal{U}$. The support set is defined as $\mathcal{L} = \{(x_i, y_i)\}_{i=1}^{N_L}$, where $x_i$ denotes an image sample and $y_i \in \{1, \dots, C\}$ represents its corresponding categorical label for $C$ distinct classes. The unlabeled pool is represented as $\mathcal{U} = \{x_i\}_{i=1}^{N_U}$, where $N_L \ll N_U$.

The primary objective is to optimize a classification function $f(\cdot)$ that minimizes the empirical risk on $\mathcal{L}$ while effectively exploiting the structural information in $\mathcal{U}$ to improve generalization on unseen test data. We achieve this by iteratively expanding $\mathcal{L}$ with high-confidence samples discovered in $\mathcal{U}$ through a recursive alignment process.

\subsection{Architectural Adaptation via LoRA}

The framework utilizes the Contrastive Language-Image Pre-training (CLIP) foundation model as the primary feature extractor \cite{clip}. CLIP consists of a visual encoder $f_\phi(\cdot)$ and a text encoder $g_\psi(\cdot)$. To bridge the domain gap between natural imagery and the spectral-spatial characteristics of satellite sensors, we use Low-Rank Adaptation (LoRA) \cite{lora}. 

\begin{figure}[htbp]
    \centering
    \includegraphics[width=0.372\textwidth]{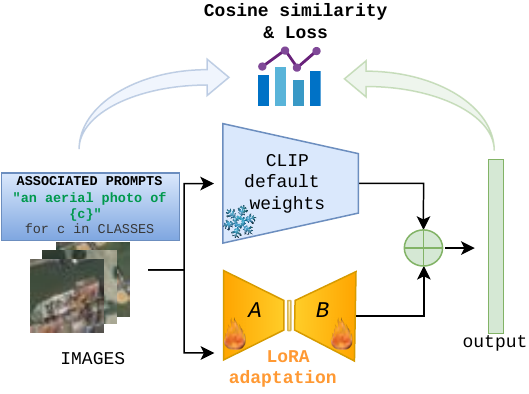}
    \caption{Overview of the Low-Rank Adaptation (LoRA) mechanism. Trainable matrices $A$ and $B$ are integrated into the frozen Transformer architecture to efficiently specialize the foundation model for remote sensing scene classification.}
    \label{fig:lora}
\end{figure}

For a given frozen weight matrix $W_0 \in \mathbb{R}^{d \times l}$ within the Transformer layers, we inject trainable bypass matrices $A \in \mathbb{R}^{r \times l}$ and $B \in \mathbb{R}^{d \times r}$ (see Fig.~\ref{fig:lora}). The updated forward pass is formulated as:
\begin{equation}
    h_i = W_0 x_i + \Delta W x_i = W_0 x_i + BAx_i
\end{equation}

where $r$ is the intrinsic rank, typically $r \ll \min(d, l)$. This constraint ensures that the adaptation process remains focused on the most relevant features while preventing overfitting to the sparse initial labels.

\vspace{-5mm}

\subsection{Semantic Anchoring and Alignment}

Rather than employing a traditional linear classification head, the framework utilizes fixed textual anchors $T_c \in \mathbb{R}^d$ for each class $c$. These anchors are generated by encoding class-specific prompts through the frozen text encoder. The visual representation of an image $x_i$ is mapped to a vector $V_i = f_{\phi+\Delta W}(x_i)$, and the model is trained to align $V_i$ with its respective anchor $T_c$.

The learning objective is defined by a cross-entropy loss over the temperature-scaled cosine similarities:
\begin{equation}
    \mathcal{L} = - \log \frac{\exp(\cos(V_i, T_{GT}) \cdot e^\tau)}{\sum_{j=1}^{C} \exp(\cos(V_i, T_j) \cdot e^\tau)}
\end{equation}
where $T_{GT}$ represents the ground-truth textual anchor, and $e^\tau$ denotes the constant log-parameterized temperature scale inherited directly from the pre-trained CLIP baseline to preserve calibrated cross-modal alignment.

\subsection{Recursive Discovery}
The training follows a recursive protocol designed to eliminate manual labeling requirements. Initially, the model undergoes a warm-up phase where the LoRA parameters are updated exclusively on the initial expert seeds to stabilize the gradients. 

Upon completion of the warm-up, the model initiates the discovery phase. For each class $c$, the system ranks all samples in $\mathcal{U}$ according to their similarity to $T_c$.
A budget of $k$ samples is then migrated to the support set and never re-evaluated:
\begin{equation}
    \mathcal{M}_c = \{ x_i \in \mathcal{U} \mid \text{sort}(\cos(V_i, T_c)) \le k \}
\end{equation}
where $\text{sort}(\cdot)$ sorts the unlabeled pool $\mathcal{U}$ in descending order based on their similarity to the text anchor $T_c$.

\begin{figure}[htbp]
    \centering
    \includegraphics[width=0.4\textwidth]{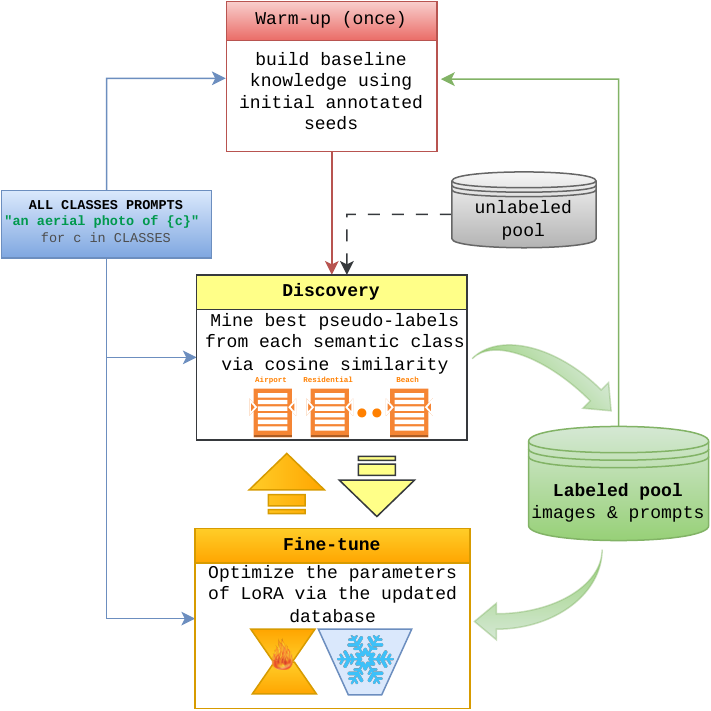}
    \caption{The SE-CLIP recursive discovery pipeline. The model iteratively synchronizes visual-textual representations on labeled seeds before mining the most similar unlabeled samples to expand the support set.}
    \label{fig:pipeline}
\end{figure}

The discovered set $\mathcal{M}_c$ is appended to $\mathcal{L}$, and the process repeats. This recursive expansion ensures a balanced and high-purity growth of the labeled pool, facilitating the continuous evolution of the adapted visual encoder. By enforcing a uniform budget $k$ across all categories, the framework ensures a strictly class-balanced expansion of the support set. This equidistribution prevents the optimization process from being dominated by high-frequency or easily discriminable classes. The flowchart of the proposed method is given in Fig.~\ref{fig:pipeline}.

\vspace{-5mm}

\section{Experiments} \label{sec:experiments}

\subsection{Datasets and Settings}
Evaluations are carried out on two benchmarks, namely UC Merced (UCM) \cite{ucm} and NWPU-RESISC45 (NWPU) \cite{nwpu}. The UCM dataset contains 2100 aerial images evenly distributed across 21 land-use categories, whereas the large-scale NWPU dataset contains 31,500 images across 45 classes (700 per class). All scene images are resized to $256 \times 256$ pixels. 
The framework is implemented in PyTorch utilizing OpenAI's pre-trained ViT-B/32 CLIP model as the core multimodal backbone. Optimization is executed via Adam with a learning rate of $10^{-4}$. 
The warm up iterations were set to 10, and the total training iterations to 200.
All the experiments have been run five times with different random seeds, and the mean $\pm$ standard deviation is reported in each case.

Following prior works regarding data division \cite{darp}, $80\%$ of the dataset is allocated for training, where only five seeds per class are initially labeled, and the remainder is treated as unlabeled data, while the remaining $20\%$ is strictly reserved for testing. Classification accuracy is adopted as a metric.

\subsection{Quantitative Analysis}
To evaluate the sensitivity of the proposed framework to the pseudo-label mining budget, Table \ref{tab:rank4_explore_k} reports the classification scores across various $k$ values with a fixed LoRA rank of $r = 4$. For the UCM dataset, the performance remains highly stable, achieving a peak accuracy of $98.81\%$ at $k = 1$ before experiencing a marginal degradation as $k$ increases. This suggests that a smaller, high-confidence mining budget is sufficient for less complex scene distributions. Conversely, on the more challenging and large-scale NWPU dataset, performance scales directly with the mining budget, rising from $88.56\%$ at $k = 1$ to a peak of $94.92\%$ at $k = 5$. This improvement underscores that complex, multi-class remote sensing environments benefit from a larger pool of mined pseudo-labels per iteration to effectively align the visual embeddings with the text anchors. However, increasing $k$ further to $7$ introduces a slight performance drop ($94.77\%$), suggesting the inclusion of noisy or misclassified samples into the training pool.

\begin{table}[htbp]
\centering
\caption{SCORES (\%) ON UCM AND NWPU DATASETS WITH RANK $r = 4$ ACROSS VARIOUS $k$ VALUES}
\label{tab:rank4_explore_k}
\begin{tabular}{lcccc}
\hline
\textbf{Dataset} & \multicolumn{4}{c}{\textbf{Top-$k$}} \\
\cline{2-5}
 & \textbf{1} & \textbf{2} & \textbf{5} & \textbf{7} \\
\hline
UCM  & \textbf{98.81 $\pm$ 0.69} & 98.62 $\pm$ 0.39 & 98.62 $\pm$ 0.49 & 98.38 $\pm$ 0.78 \\
NWPU & 88.56 $\pm$ 1.32          & 91.52 $\pm$ 0.48          & \textbf{94.92 $\pm$ 0.31} & 94.77 $\pm$ 0.48 \\
\hline
\end{tabular}
\end{table}

In the second experiment, Table \ref{tab:explore_lora_rank} evaluates the impact of the LoRA rank $r$ under the values established previously ($k=1$ for UCM and $k=5$ for NWPU). The results demonstrate a monotonic performance improvement on both datasets as the rank scales. Specifically, increasing $r$ from 2 to 16 improves the classification accuracy from $98.71\%$ to $99.09\%$ on UCM, and from $94.41\%$ to $95.07\%$ on NWPU. This slight gain indicates that a higher low-rank parameter space provides more learning capacity to adapt the underlying visual encoder to the complex spatial textures and highly correlated class boundaries inherent in remote sensing imagery.

The ablation results presented in Table \ref{tab:ablation_recursive_vs_warmup} validate the necessity of the full recursive discovery mechanism. Relying solely on the warm-up phase causes significant performance drops on both benchmarks. This confirms that, while the warm-up initialization provides a stable starting point, continuous recursive optimization is essential for the model to successfully discover and assimilate new samples without collapsing.

The evaluation in Table \ref{tab:ablation_balanced_vs_global} (for $k=5$ and $r=4$) highlights the critical role of our per-class balanced mining strategy over a standard global mining baseline (i.e., aggregating all the samples and selecting the global top ones without per-class constraints). Without class-specific constraints, global mining suffers catastrophic performance degradation on both datasets, alongside elevated standard deviations. Enforcing local per-class balance completely mitigates this instability, yielding superior accuracy and verifying that uniform category distribution is essential for robust semi-supervised adaptation.

\begin{figure}[htbp]
    \centering
    \includegraphics[width=0.5\textwidth]{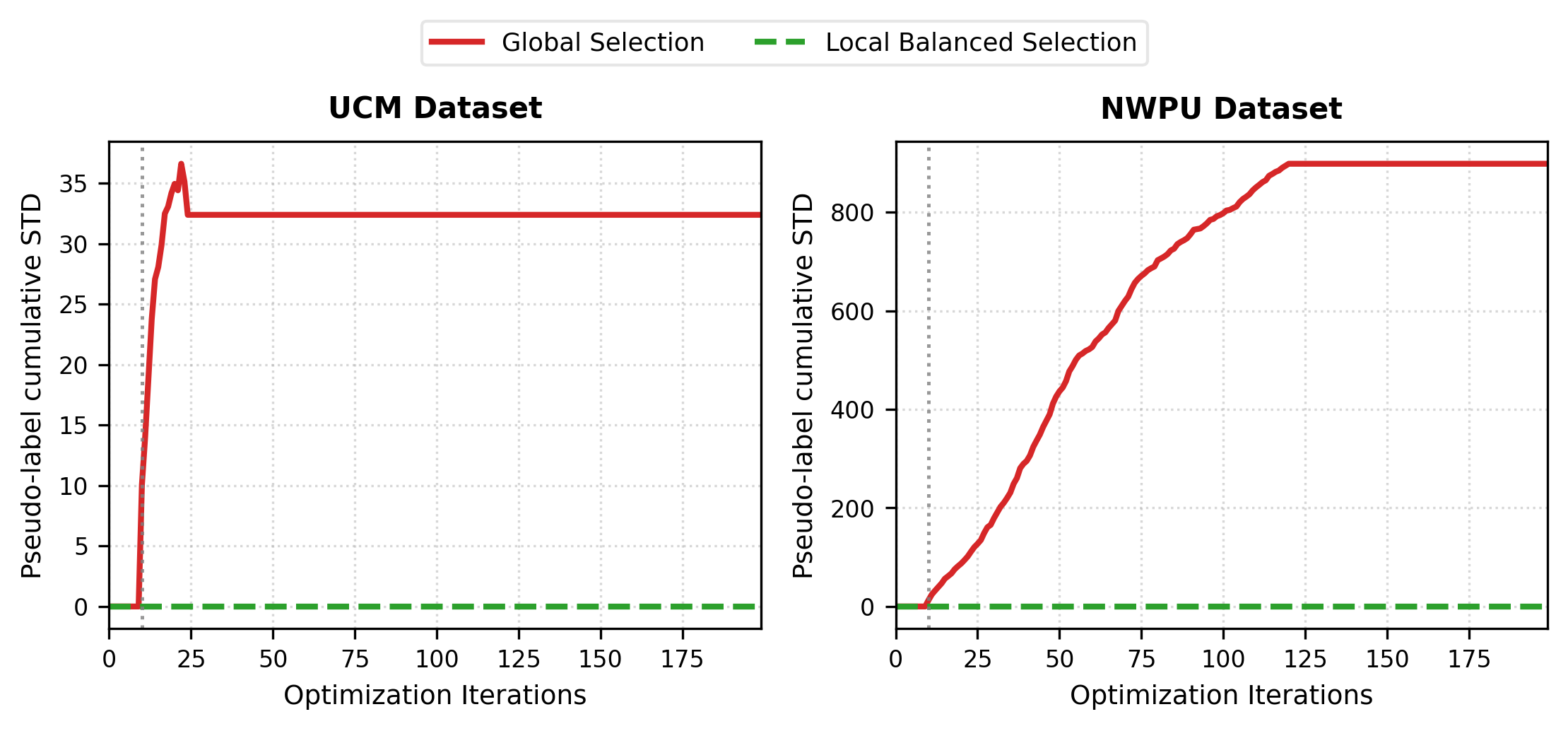}
    \caption{Cumulative pseudo-label standard deviation across iterations for global vs. local selection on UCM and NWPU.}
    \label{fig:class_imbalance_evolution}
\end{figure}

To further study the effect of the proposed balanced per-class mining mechanism, we calculate the cumulative standard deviation in comparison to using a standard global mining baseline. As illustrated in Fig.~\ref{fig:class_imbalance_evolution}, the global selection baseline exhibits an exponential surge in cumulative standard deviation immediately following the 10-iteration warm-up phase, suggesting easy class dominance. Conversely, our local selection strategy maintains a strict standard deviation of zero across all optimization iterations on both datasets, providing empirical evidence that enforcing per-class constraints successfully stabilizes pseudo-label distribution.

\begin{table}[htbp]
\centering
\caption{SCORES (\%) ON UCM AND NWPU DATASETS ACROSS VARIOUS LoRA RANK ($r$) CONFIGURATIONS}
\label{tab:explore_lora_rank}
\begin{tabular}{lccc}
\hline
\textbf{Dataset} & \multicolumn{3}{c}{\textbf{LoRA Rank ($r$)}} \\
\cline{2-4}
 & \textbf{2} & \textbf{4} & \textbf{16} \\
\hline
UCM  & 98.71 $\pm$ 0.27 & 98.81 $\pm$ 0.69 & \textbf{99.09 $\pm$ 0.54} \\
NWPU & 94.41 $\pm$ 0.34 & 94.92 $\pm$ 0.31 & \textbf{95.07 $\pm$ 0.95} \\
\hline
\end{tabular}
\end{table}

\begin{table}[htbp]
\centering
\caption{ABLATION STUDY COMPARING THE FULL RECURSIVE DISCOVERY MECHANISM AGAINST WARM-UP ONLY}
\label{tab:ablation_recursive_vs_warmup}
\begin{tabular}{lcc}
\hline
\textbf{Configuration} & \textbf{UCM} & \textbf{NWPU} \\
\hline
Full recursive discovery & \textbf{99.09 $\pm$ 0.54} & \textbf{95.07 $\pm$ 0.95} \\
Warm-up only             & 87.81 $\pm$ 2.50          & 76.34 $\pm$ 0.87          \\
\hline
\end{tabular}
\end{table}

\begin{table}[htbp]
\centering
\caption{ABLATION STUDY COMPARING PER-CLASS BALANCED MINING AGAINST GLOBAL MINING}
\label{tab:ablation_balanced_vs_global}
\begin{tabular}{lcc}
\hline
\textbf{Configuration} & \textbf{UCM} & \textbf{NWPU} \\
\hline
Per-class balanced mining & \textbf{98.62 $\pm$ 0.49} & \textbf{94.92 $\pm$ 0.31} \\
Global mining             & 83.19 $\pm$ 2.55          & 42.33 $\pm$ 4.69          \\
\hline
\end{tabular}
\end{table}

\begin{table}[htbp]
\centering

\caption{COMPUTATIONAL EFFICIENCY ANALYSIS}
\label{tab:computational_efficiency}
\resizebox{\columnwidth}{!}{%
\begin{tabular}{ll}
\hline
\textbf{Metric Description} & \textbf{Value / Footprint} \\ \hline
Hardware environment & NVIDIA Tesla V100 GPU (32 GB) \\
Total parameters & 152.54 M \\
Trainable parameters ($r=16$) & 1,769,472 (1.16\%) \\
 GPU memory consumption & 
 1.12 GB (batch size = 16) \\
 FLOPs & 
 2.95 GFLOPs/image \\
UCM total training time & 26,933.52 sec (7.48 hours) \\
NWPU total training time & 52,649.98 sec (14.62 hours) \\
Inference latency (per batch, size=16) & 10.26 ms \\ \hline
\end{tabular}%
}
\end{table}

\begin{figure}[htbp]
    \centering
    \includegraphics[width=0.3\textwidth]{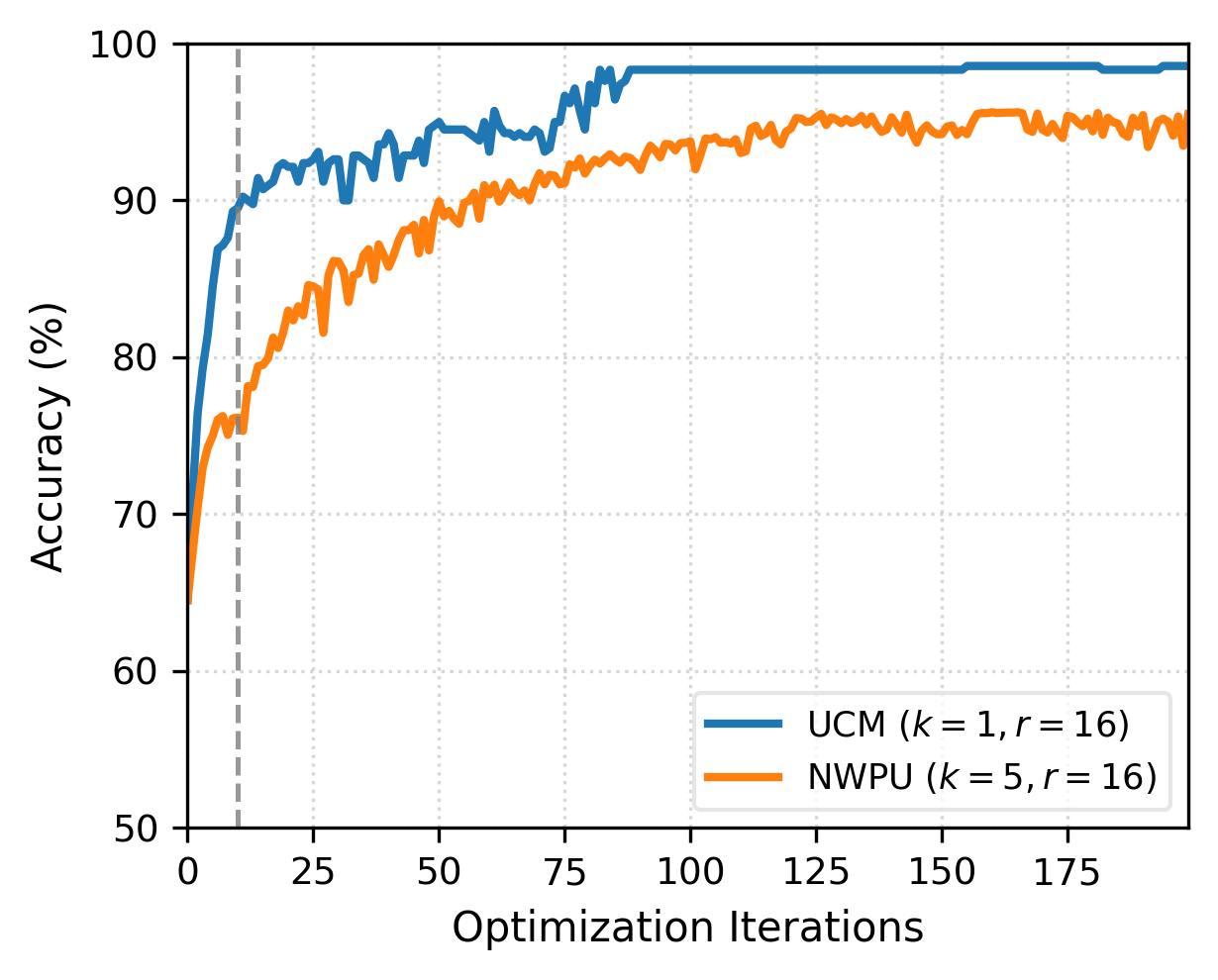}
    \caption{Accuracy (\%) evolution of SE-CLIP across 200 iterations on the UCM and NWPU benchmarks. The grey dashed vertical line marks the mining onset.}

    \label{fig:acc_evolution}
\end{figure}

\begin{figure*}[htbp]
    \centering
    \vspace{-10mm}\includegraphics[width=0.6\textwidth]{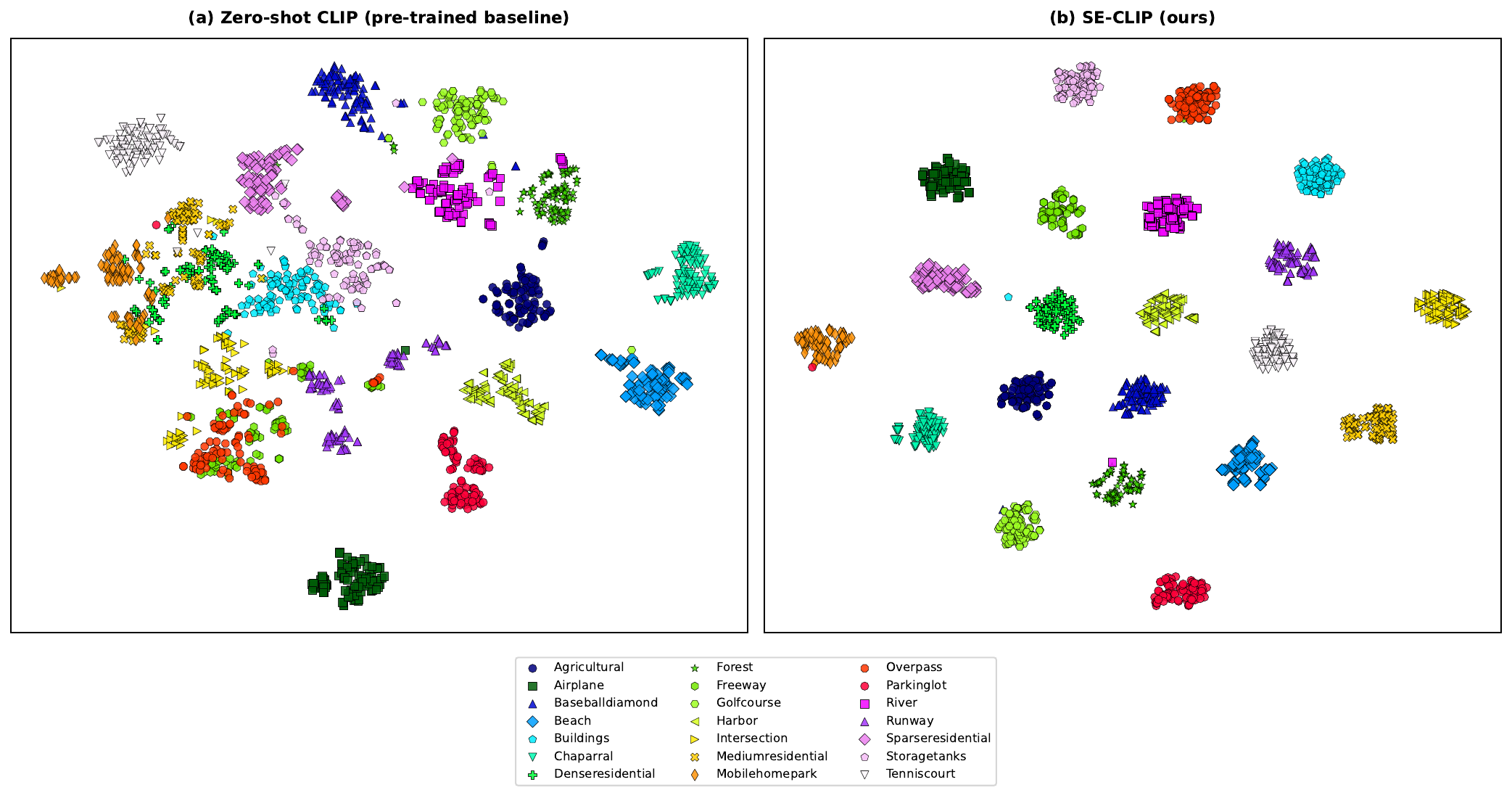}
    \includegraphics[width=0.6\textwidth]{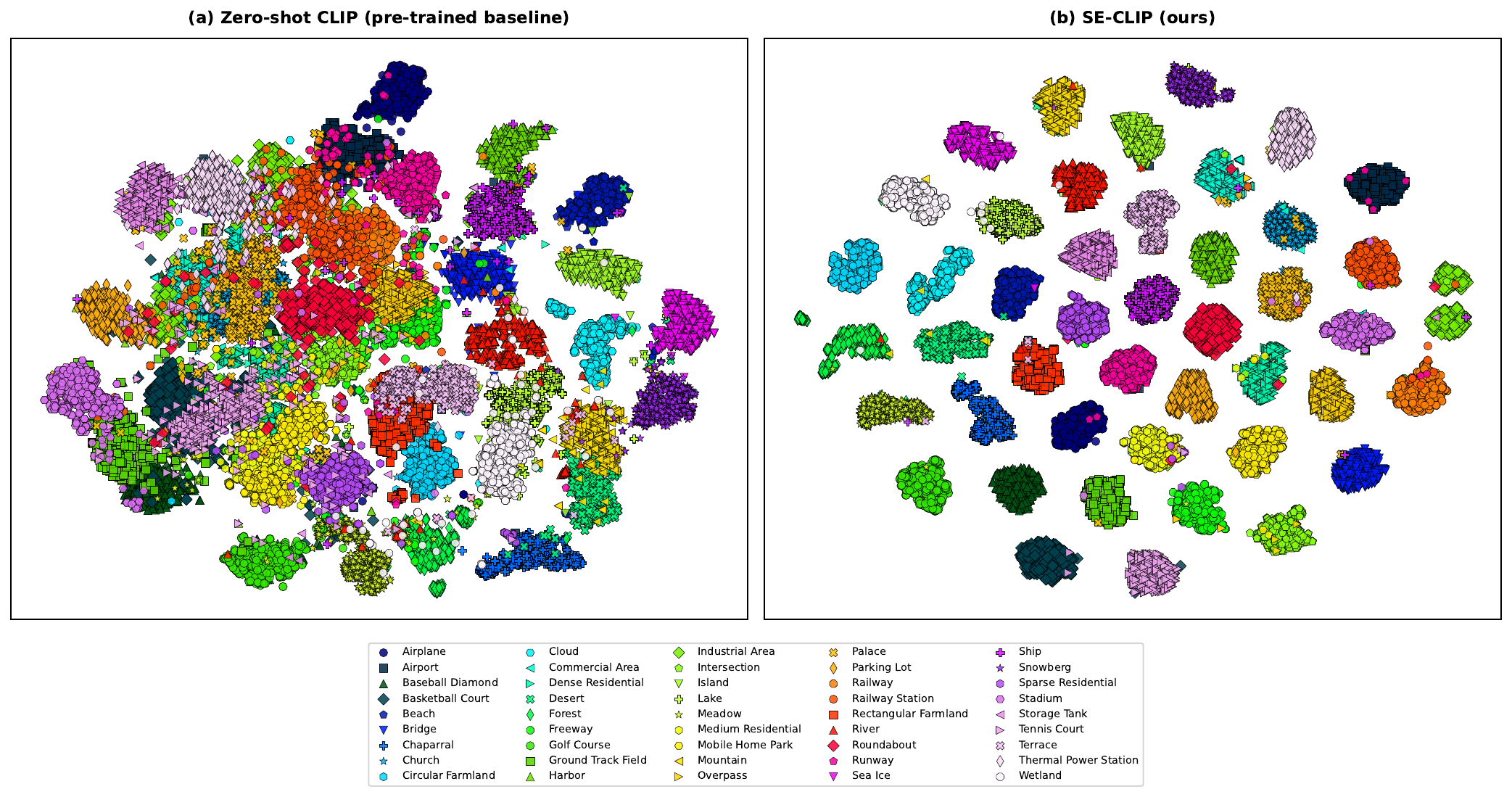}

    \caption{Qualitative t-SNE visualization \cite{tsne} of the visual embedding space comparing the pre-trained zero-shot CLIP baseline (left panels) against our proposed SE-CLIP method (right panels) across all categories for both UCM (top) and NWPU (bottom) datasets.
    The proposed SE-CLIP successfully adapts and reorganizes the visual embeddings into tightly bound, well-separated semantic clusters.
    Distinct marker shapes and color schemes are assigned uniquely to each category to highlight cluster separability. 
    Best viewed in color.} 

    \label{fig:tsne_qualitative}
\end{figure*}

\begin{table}[htbp]
\centering
\caption{COMPARISON WITH STATE-OF-THE-ART SEMI-SUPERVISED METHODS}
\label{tab:sota_comparison}
\begin{tabular}{lccc}
\hline
\textbf{Method} & \textbf{UCM (\%)} & \textbf{NWPU (\%)} \\
\hline
Flexmatch \cite{flexmatch} & 82.54 & 69.16 \\
MSmatch \cite{msmatch} & 85.70 & 60.00 \\
RSmatch \cite{rsmatch} & 84.10 & 58.30 \\
Freematch \cite{freematch}  & 80.71 & 69.10 \\
Softmatch \cite{softmatch} & 87.38 & 70.09 \\
Fixmatch \cite{fixmatch}   & 84.84 & 65.89 \\
SemiRS-COC \cite{semirs}& 85.00 & 59.10 \\
CPL-PL \cite{cpl}    & 87.20 & 61.30 \\
DARP \cite{darp}     & 95.87 & 88.86 \\
\hline
\rowcolor{gray!20} SE-CLIP (OURS) & \textbf{99.09} & \textbf{95.07} \\
\hline
\end{tabular}
\end{table}

We further analyze the computational efficiency of SE-CLIP. As detailed in Table~\ref{tab:computational_efficiency}, our framework optimizes only 1.16\% of the CLIP model volume via LoRA. Moreover, because the recursive label-mining modules are completely discarded during deployment, our framework achieves a rapid inference latency of just 10.26~ms per batch (size=16).

\subsection{Qualitative analysis}
The performance trajectories across optimization iterations are shown in Fig. \ref{fig:acc_evolution}. Following the 10-iteration warm-up, the activation of top-$k$ label mining triggers a smooth accuracy ascent on both benchmarks. Due to varying dataset scale and complexity, the performance stabilizes into a plateau around iteration 60 for UCM, whereas the larger NWPU dataset requires a prolonged discovery trajectory to reach its steady-state plateau between iterations 160 and 170. As illustrated by the t-SNE visualizations \cite{tsne} in Fig. \ref{fig:tsne_qualitative} the zero-shot CLIP baseline exhibits highly overlapping class distributions with poorly defined boundaries. By contrast, our proposed SE-CLIP successfully reorganizes the visual embedding space into tightly bound, well-separated semantic clusters, confirming its superior capability in maximizing inter-class separability.

To verify the competitive advantage of the proposed framework, Table \ref{tab:sota_comparison} compares SE-CLIP against several state-of-the-art semi-supervised learning methods on both benchmarks. The results show that SE-CLIP outperforms representative state-of-the-art semi-supervised frameworks. Specifically, on the UCM dataset, our method achieves an accuracy of $99.09\%$, outperforming the strongest baseline, DARP \cite{darp}, by $3.22\%$. On the more challenging NWPU benchmark,  SE-CLIP reaches $95.07\%$, with a substantial $6.21\%$ improvement over the strongest baseline. blue The results confirm that using CLIP with parameter-efficient low-rank adaptation provides superior bimodal representation capabilities in few-shot remote sensing scenarios compared to standard semi-supervised frameworks.
\vspace{-5mm}

\section{Conclusion} \label{sec:conclusion}

In this letter, we presented SE-CLIP, a semi-supervised scene classification framework that leverages the pre-trained capabilities of the CLIP foundation model through parameter-efficient fine-tuning. This is achieved by applying a balanced pseudo-label mining strategy based on fixed textual anchors. The experiments show that adapting CLIP, pretrained on natural images, via target-directed low-rank mining offers a more effective alternative for handling complex remote sensing data distributions than traditional unimodal classifiers.

Regarding operational constraints, the framework requires expressive text vocabularies to construct discriminative anchors, and performance remains limited in fine-grained scenarios with high inter-class structural similarity where semantic descriptions cannot fully resolve visual overlap. To this end, future work will focus on developing robust VLM-driven paradigms to improve cross-modal alignment for image classification using only class names \cite{orion}.

\vspace{-5mm}

\section*{Acknowledgment}
This research was supported by Ongoing Research Funding program (ORF-2026-995), King Saud University, Riyadh, Saudi Arabia.
\vspace{-5mm}
\ifCLASSOPTIONcaptionsoff
  \newpage
\fi

\bibliographystyle{IEEEtran}
\bibliography{references}

@inproceedings{clip,
  title={Learning Transferable Visual Models from Natural Language Supervision},
  author={Radford, Alec and Kim, Jong Wook and Hallacy, Chris and Ramesh, Aditya and Goh, Gabriel and Agarwal, Sandhini and Sastry, Girish and Askell, Amanda and Mishkin, Pamela and Clark, Jack and others},
  booktitle={International Conference on Machine Learning},
  pages={8748--8763},
  year={2021},
  organization={PMLR}
}

@article{li2026clip,
  title={Clip-powered domain generalization and domain adaptation: A comprehensive survey},
  author={Li, Jindong and Li, Yongguang and Fu, Yali and Liu, Jiahong and Liu, Yixin and Yang, Menglin and King, Irwin},
  journal={IEEE Transactions on Pattern Analysis and Machine Intelligence},
  year={2026},
  publisher={IEEE}
}

@article{tsne,
  title={Visualizing data using t-SNE},
  author={Van der Maaten, Laurens and Hinton, Geoffrey},
  journal={Journal of Machine Learning Research},
  volume={9},
  number={11},
  pages={2579--2605},
  year={2008}
}

@article{flexmatch,
  title={Flexmatch: Boosting semi-supervised learning with curriculum pseudo labeling},
  author={Zhang, Bowen and Wang, Yidong and Hou, Wenxin and Wu, Hao and Wang, Jindong and Okumura, Manabu and Shinozaki, Takahiro},
  journal={Advances in neural information processing systems},
  volume={34},
  pages={18408--18419},
  year={2021}
}

@article{msmatch,
  title={MSMatch: Semisupervised multispectral scene classification with few labels},
  author={G{\'o}mez, Pablo and Meoni, Gabriele},
  journal={IEEE Journal of Selected Topics in Applied Earth Observations and Remote Sensing},
  volume={14},
  pages={11643--11654},
  year={2021},
  publisher={IEEE}
}

@inproceedings{rsmatch,
  title={RSMatch: Semi-supervised learning with adaptive category-related pseudo labeling for remote sensing scene classification},
  author={Lin, Weiquan and Ma, Jingjing and Tang, Xu and Zhang, Xiangrong and Jiao, Licheng},
  booktitle={International Conference on Intelligence Science},
  pages={220--227},
  year={2022},
  organization={Springer}
}

@article{freematch,
  title={Freematch: Self-adaptive thresholding for semi-supervised learning},
  author={Wang, Yidong and Chen, Hao and Heng, Qiang and Hou, Wenxin and Fan, Yue and Wu, Zhen and Wang, Jindong and Savvides, Marios and Shinozaki, Takahiro and Raj, Bhiksha and others},
  journal={arXiv preprint arXiv:2205.07246},
  year={2022}
}

@article{softmatch,
  title={Softmatch: Addressing the quantity-quality trade-off in semi-supervised learning},
  author={Chen, Hao and Tao, Ran and Fan, Yue and Wang, Yidong and Wang, Jindong and Schiele, Bernt and Xie, Xing and Raj, Bhiksha and Savvides, Marios},
  journal={arXiv preprint arXiv:2301.10921},
  year={2023}
}

@inproceedings{fixmatch,
  title={Self adaptive threshold pseudo-labeling and unreliable sample contrastive loss for semi-supervised image classification},
  author={Zhang, Xuerong and Huang, Li and Lv, Jing and Yang, Ming},
  booktitle={International Conference on Artificial Neural Networks},
  pages={61--75},
  year={2024},
  organization={Springer}
}

@article{semirs,
  title={SemiRS-COC: Semi-supervised classification for complex remote sensing scenes with cross-object consistency},
  author={Liu, Qiang and Yue, Jun and Kuang, Yang and Xie, Weiying and Fang, Leyuan},
  journal={IEEE Transactions on Image Processing},
  volume={33},
  pages={3855--3870},
  year={2024},
  publisher={IEEE}
}

@article{cpl,
  title={CPL-PL: Contrapositive Learning-Based Pseudo-Labeling for Semi-Supervised Scene Classification in Remote Sensing Images},
  author={Swetha, G and Datla, Rajeshreddy and Babu, Sobhan and Mohan, C Krishna},
  journal={IEEE Geoscience and Remote Sensing Letters},
  year={2025},
  publisher={IEEE}
}

@article{darp,
  title={Semi-supervised Remote Sensing Imagery Scene Classification Based on Probabilistic Selection},
  author={Ye, Lixin and Shang, Fanhua and Liu, Hongying},
  journal={IEEE Geoscience and Remote Sensing Letters},
  year={2025},
  publisher={IEEE}
}

@inproceedings{ucm,
  title={Bag-of-visual-words and spatial extensions for land-use classification},
  author={Yang, Yi and Newsam, Shawn},
  booktitle={Proceedings of the 18th ACM SIGSPATIAL International Conference on Advances in Geographic Information Systems},
  pages={270--279},
  year={2010}
}

@article{nwpu,
  title={Remote sensing image scene classification: Benchmark and state of the art},
  author={Cheng, Gong and Han, Junwei and Lu, Xiaoqiang},
  journal={Proceedings of the IEEE},
  volume={105},
  number={10},
  pages={1865--1883},
  year={2017},
  publisher={IEEE}
}

@inproceedings{orion,
  title={ORION: ORthonormal Text Encoding for Universal VLM AdaptatION},
  author={Chakraborty, Omprakash and Dolz, Jose and Ben Ayed, Ismail},
  booktitle={Proceedings of the IEEE/CVF Conference on Computer Vision and Pattern Recognition},
  pages={31556--31565},
  year={2026}
}

@article{lora,
  title={Lora: Low-rank adaptation of large language models.},
  author={Hu, Edward J and Shen, Yelong and Wallis, Phillip and Allen-Zhu, Zeyuan and Li, Yuanzhi and Wang, Shean and Wang, Liang and Chen, Weizhu and others},
  journal={Iclr},
  volume={1},
  number={2},
  pages={3},
  year={2022}
}

% % --- Clear the page layout to prevent style bleeding ---
% \clearpage 
% % --- Append the entire Response to Reviewers PDF ---
% % This command imports every page of your response PDF, keeping its original single-column layout, margins, and fonts perfectly intact.
% \includepdf[pages=-]{Response to the Reviewers’ Comments.pdf}

\end{document}